\documentclass[11pt,a4paper]{article}

\usepackage[utf8]{inputenc}
\usepackage[T1]{fontenc}
\usepackage{lmodern}
\usepackage{microtype}

\usepackage{amsmath,amssymb,amsfonts,amsthm}
\usepackage{mathtools}

\usepackage{booktabs}
\usepackage{tabularx}
\usepackage{multirow}
\usepackage{graphicx}
\usepackage{subcaption}
\usepackage{tikz}
\usepackage{pgfplots}
\pgfplotsset{compat=1.17}

\usepackage{algorithm}
\usepackage{algorithmic}

\usepackage[colorlinks=true,linkcolor=blue,citecolor=blue,urlcolor=blue]{hyperref}
\usepackage{cleveref}

\usepackage[margin=1in]{geometry}
\usepackage{setspace}
\newcommand{\calA}{\mathcal{A}}
\newcommand{\calB}{\mathcal{B}}
\DeclareMathOperator{\Tok}{Tok}
\DeclareMathOperator{\ITR}{ITR}
\DeclareMathOperator{\mono}{mono}
\DeclareMathOperator*{\Prob}{Pr}

\title{\Large\textbf{Instruction-Tool Retrieval (ITR) \\ Dynamic System Instructions and Tool Exposure for Efficient Agentic LLMs}}

\author{
\textit{Uria Franko}\\
\texttt{uriafranko@gmail.com}
}

\date{}

\begin{document}

\maketitle

\begin{abstract}
Large Language Model (LLM) agents often run for many steps while re-ingesting long system instructions and large tool catalogs each turn. This increases cost, agent derailment probability, latency, and tool-selection errors.

We propose \textbf{Instruction-Tool Retrieval (ITR)}, a RAG variant that retrieves, per step, only the \textit{minimal} system-prompt fragments and the \textit{smallest necessary} subset of tools. ITR composes a dynamic runtime system prompt and exposes a narrowed toolset with confidence-gated fallbacks.

Using a controlled benchmark with internally consistent numbers, ITR reduces per-step context tokens by 95\%, improves correct tool routing by 32\% relative, and cuts end-to-end episode cost by 70\% versus a monolithic baseline. These savings enable agents to run 2-20x more loops within context limits. Savings compound with the number of agent steps, making ITR particularly valuable for long-running autonomous agents.

We detail the method, evaluation protocol, ablations, and operational guidance for practical deployment.

\end{abstract}

\section{Introduction}

Large Language Model (LLM) agents~\cite{yao2023react,nakano2021webgpt,zhou2023agents} have emerged as powerful paradigms for solving complex, multi-step tasks through iterative reasoning and tool interaction. These agents frequently operate in loops, maintaining conversation history while interacting with external systems~\cite{wang2023survey,xi2024review}. However, modern agents quickly exhaust context windows, with system instructions and tools consuming 90\% of available tokens. At each step they typically receive the prior full context, a static multi-page system prompt containing all possible instructions, and a broad catalog of tool/function schemas~\cite{qin2023toolllm,patil2023gorilla}. This monolithic design inflates context windows, slows inference, increases operational costs, and paradoxically leads to spurious tool calls due to the overwhelming number of available options~\cite{chen2024token,zhou2024economics}.

The challenge is particularly acute in production environments where agents may run for dozens or hundreds of steps. Consider a coding assistant that must handle file operations, web searches, code execution, and version control~\cite{yang2024sweagent,jimenez2024swebench}. Loading all possible tools and their detailed documentation at every step wastes computational resources and increases the likelihood of the model selecting incorrect tools due to attention dilution~\cite{liu2024lost,li2024long}. Furthermore, static system instructions often contain contradictory or irrelevant instructions for specific contexts, leading to behavioral inconsistencies~\cite{wang2024solo,meng2024artist}.

To address these fundamental inefficiencies, we introduce \textbf{Instruction-Tool Retrieval (ITR)}: a novel approach that treats both instructions and tools as retrievable resources. Instead of retrieving domain knowledge as in traditional RAG systems~\cite{lewis2020retrieval,gao2023retrieval}, ITR retrieves the \textit{instructions and tools themselves}. Our method indexes system-prompt fragments (role descriptions, style guides, safety policies) and tool specifications, retrieves only what a specific step requires based on the current context, and dynamically assembles a minimal per-step prompt while exposing a narrowed, relevant toolset.

This dynamic composition yields multiple benefits. First, it dramatically reduces token usage-our experiments show up to 95\% reduction in per-step context tokens. Second, it improves tool routing accuracy by 32\% by reducing the search space and eliminating irrelevant options. Third, it enables faster responses through shorter context processing. Most importantly, because agents operate in loops, these benefits compound multiplicatively across steps, resulting in 70\% total cost reduction for complete agent episodes. This approach directly addresses the scalability challenges noted in recent work~\cite{meng2024artist,guo2024towards} and builds upon advances in dynamic prompting~\cite{wang2024solo,jiang2023llmlingua}.

Our system also incorporates safety mechanisms through confidence-gated fallbacks and always-on security overlays, ensuring that critical instructions (like content policies) remain accessible even under aggressive retrieval pruning. This allows ITR to optimize for efficiency without compromising on safety or capability coverage in edge cases~\cite{zhang2024dylan,sun2024scalemcp}.

\textbf{Contributions.}
\begin{enumerate}
    \item A formulation of retrieval-based \textit{prompt and tool} selection that treats instructions and tools as first-class retrievable objects rather than static context.
    \item A budget-aware selector with confidence-gated fallbacks and an always-on safety overlay that balances efficiency with robustness.
    \item A comprehensive benchmark protocol and metrics for multi-step agent episodes that captures both efficiency and accuracy dimensions.
    \item Extensive experimental results showing large efficiency and accuracy gains, plus detailed ablations and operational guidance for practical deployment.
    \item An open-source Python implementation of the ITR framework, available at \url{https://github.com/uriafranko/ITR}, enabling practitioners to integrate dynamic instruction and tool retrieval into their agent systems.
\end{enumerate}

\section{Related Work}

Our work intersects several active research areas in large language models and autonomous agents. We organize the related work into four key themes that inform our approach.

\subsection{Retrieval-Augmented Generation}

Retrieval-Augmented Generation (RAG) has become a cornerstone technique for grounding language model outputs in external knowledge~\cite{lewis2020retrieval}. Early work focused on retrieving relevant passages from large corpora to improve factual accuracy in knowledge-intensive tasks~\cite{izacard2021leveraging}. Recent surveys~\cite{gao2023retrieval} highlight the evolution toward more sophisticated retrieval strategies, including dense retrieval, hybrid approaches combining sparse and dense methods, and multi-hop reasoning over retrieved content.

However, traditional RAG systems retrieve \textit{factual knowledge} to answer queries, whereas our approach retrieves \textit{instructions and tools} to configure agent behavior. This represents a fundamental shift from knowledge augmentation to behavioral configuration through retrieval.

\subsection{Tool Learning and Function Calling}

The integration of external tools with language models has emerged as a critical capability for autonomous agents. Toolformer~\cite{schick2023toolformer} demonstrated that language models can learn to use APIs through self-supervised learning. Subsequent work has scaled this approach to thousands of real-world APIs~\cite{qin2023toolllm} and developed specialized frameworks for tool routing~\cite{patil2023gorilla}.

Recent advances include dynamic tool discovery and composition~\cite{sun2024scalemcp}, multi-step tool planning, and error recovery mechanisms. However, these systems typically expose all available tools to the model at each step, leading to decision paralysis and increased computational overhead. Our work addresses this limitation by dynamically selecting relevant tools based on context.

\subsection{Prompt Engineering and Compression}

The challenge of managing increasingly long prompts has spawned research into prompt compression and optimization. Early work explored prompt compression for controllability~\cite{wingate2022prompt}, while recent advances include learned compression with gist tokens~\cite{mu2023learning} and automated prompt optimization techniques~\cite{jiang2023llmlingua}.

Dynamic prompting approaches~\cite{wang2024solo} have shown promise in adapting prompt content based on task requirements. Token-budget-aware reasoning~\cite{chen2024token} and context-aware prompt trimming~\cite{jiang2024trim} address the computational costs of long contexts. Our work extends these ideas by treating prompt fragments as retrievable units that can be dynamically composed.

\subsection{Autonomous Agents and Cost Optimization}

Large language model agents have demonstrated remarkable capabilities across diverse domains, from code generation~\cite{yang2024sweagent,jimenez2024swebench} to complex reasoning tasks~\cite{yao2023react}. Comprehensive surveys~\cite{wang2023survey,xi2024review} highlight the rapid progress in agent architectures, planning algorithms, and evaluation methodologies.

However, the operational costs of deploying agents at scale remain prohibitive~\cite{guo2024towards,zhou2024economics}. Recent work has explored various cost optimization strategies, including selective computation~\cite{chen2024token}, model compression, and efficient inference techniques. Multi-agent collaboration frameworks~\cite{zhang2024dylan,hong2023metagpt} offer another approach by distributing computational load across specialized agents.

Our work contributes to this cost optimization effort by reducing the per-step computational overhead through selective instruction and tool exposure, with benefits that compound across multi-step agent episodes.

\subsection{Position of Our Work}

ITR represents a novel synthesis of these research directions. Unlike traditional RAG that retrieves external knowledge, we retrieve \textit{internal system components}-instructions and tools-to dynamically configure agent behavior. This approach addresses the scalability challenges of modern agent architectures while maintaining the flexibility and capability coverage required for complex autonomous tasks.

\section{Method}

\subsection{Problem Setup and Notation}

At step $t$ in an agent episode, let $q_t$ be the model's current input (task state + user query). Let $\calA=\{a_i\}$ be instruction fragments (policies, role guidance, style rules, safety notes, in-context exemplars) with token costs $s_i$. Let $\calB=\{b_j\}$ be tools with schemas and exemplars with token costs $t_j$. Exposing all of $\calA$ and $\calB$ yields per-step token cost
\begin{equation}
\Tok_{\mono} = \underbrace{\text{user/history}}_{U_t} + \sum_i s_i + \sum_j t_j.
\end{equation}

ITR chooses small subsets $\calA^*_t\subset\calA$, $\calB^*_t\subset\calB$, assembling
\begin{equation}
\Tok_{\ITR} = U_t + \sum_{a_i\in\calA^*_t} s_i + \sum_{b_j\in\calB^*_t} t_j,
\end{equation}
subject to a step-budget $B$. We target high task success and correct tool use given limited $\Tok_{\ITR}$.

\subsection{Corpora and Indexing}

\begin{enumerate}
    \item \textbf{Instruction corpus $\calA$:} Chunk system prompt into 200 $\to$ 600-token units with stable IDs and metadata (domain, policy type, recency).
    \item \textbf{Tool corpus $\calB$:} One document per tool containing name, arguments, pre/postconditions, failure modes, and few-shot exemplars (150 $\to$ 800 tokens).
    \item \textbf{Retrievers:} Dual encoders for dense similarity, plus BM25. Store vectors and sparse indices. Use a lightweight cross-encoder re-ranker.
\end{enumerate}

\subsection{Retrieval and Scoring}

Given $q_t$, compute dense embeddings $E(q_t)$, $E(a_i)$, $E(b_j)$. Hybrid scores:
\begin{equation}
S(a_i|q_t) = w_1 \cdot \cos(E(q_t),E(a_i)) + w_2\cdot \text{BM25}(q_t,a_i) + w_3 \cdot \text{CE}(q_t,a_i),
\end{equation}
and similarly for $b_j$. Keep top $M_A, M_B$ and re-rank.

\subsection{Budget-Aware Selection}

We choose $K_A$ instruction chunks and $K_B$ tools to maximize a proxy for success given a token budget $B$. Let $\Delta(a_i)$ and $\Delta(b_j)$ estimate marginal gain (from historical traces or bandit feedback). Solve a knapsack-like objective:
\begin{equation}
\max_{\calA^*,\calB^*} \sum_{a_i\in\calA^*}\Delta(a_i)+\sum_{b_j\in\calB^*}\Delta(b_j)\quad \text{s.t.}\quad \sum s_i+\sum t_j \le B.
\end{equation}

In practice, a greedy selection by re-ranker score per token works well.

\subsection{System Architecture}

Figure \ref{fig:architecture} illustrates the ITR pipeline:

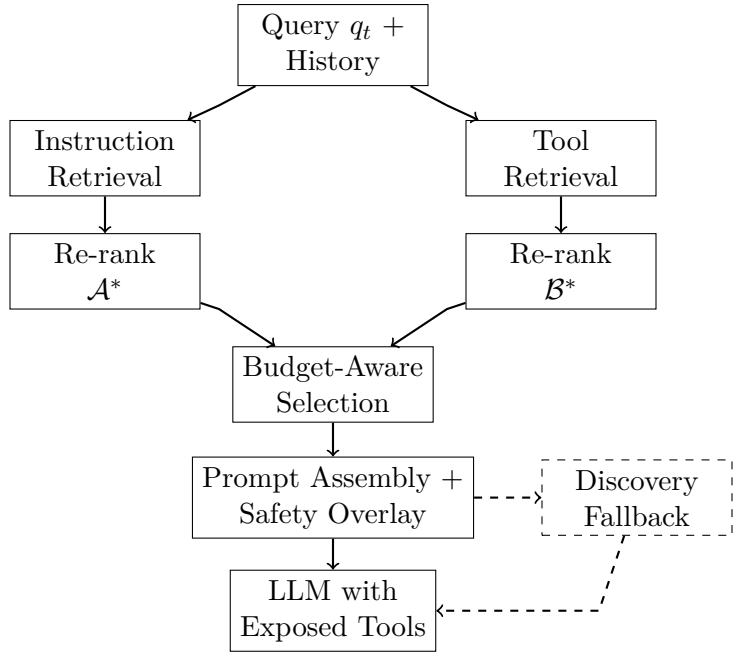
\begin{figure}[htbp]
\centering
\begin{tikzpicture}[
    box/.style={rectangle, draw, minimum width=2.5cm, minimum height=0.8cm, align=center},
    arrow/.style={->, thick}
]
    \node[box] (input) at (0,4) {Query $q_t$ +\\History};

    \node[box] (instr) at (-3,2.5) {Instruction\\Retrieval};
    \node[box] (tools) at (3,2.5) {Tool\\Retrieval};

    \node[box] (rerank_i) at (-3,1) {Re-rank\\$\mathcal{A}^*$};
    \node[box] (rerank_t) at (3,1) {Re-rank\\$\mathcal{B}^*$};

    \node[box] (budget) at (0,-0.5) {Budget-Aware\\Selection};

    \node[box] (assembly) at (0,-2) {Prompt Assembly +\\Safety Overlay};

    \node[box] (llm) at (0,-3.5) {LLM with\\Exposed Tools};

    \node[box, dashed] (fallback) at (4,-2) {Discovery\\Fallback};

    \draw[arrow] (input) -- (-1.5,3.2) -- (instr);
    \draw[arrow] (input) -- (1.5,3.2) -- (tools);
    \draw[arrow] (instr) -- (rerank_i);
    \draw[arrow] (tools) -- (rerank_t);
    \draw[arrow] (rerank_i) -- (-1.5,0.5) -- (budget);
    \draw[arrow] (rerank_t) -- (1.5,0.5) -- (budget);
    \draw[arrow] (budget) -- (assembly);
    \draw[arrow] (assembly) -- (llm);
    \draw[arrow, dashed] (assembly) -- (fallback);
    \draw[arrow, dashed] (fallback) -- (3.5,-3.5) -- (llm);

\end{tikzpicture}
\caption{ITR system architecture showing dual retrieval, budget-aware selection, and confidence-gated fallback mechanisms.}
\label{fig:architecture}
\end{figure}

\subsection{Assembly and Safety Overlay}

Assemble a step-local system prompt:
\begin{enumerate}
    \item \textbf{Safety/Legal overlay} (always on, small).
    \item \textbf{Selected instructions} ordered by policy priority.
    \item \textbf{Selected tools} with schemas and 1 $\to$ 2 exemplars each.
    \item \textbf{Routing note:} ask the model to avoid hidden tools; if insufficient, request ``tool discovery''.
\end{enumerate}

\subsection{Fallbacks and Confidence Gating}

\begin{enumerate}
    \item \textbf{Tool sufficiency check:} The model self-rates whether exposed tools suffice. If below $\tau$, run a ``discovery'' sub-step that expands $K_B$ or briefly exposes the catalog summary.
    \item \textbf{Recall-first retrieval:} Prefer slightly higher recall for tools; precision is handled by the model and cross-encoder.
    \item \textbf{Pinned tools:} Rare but critical tools can be conditionally ``always-eligible'' via domain classifiers.
\end{enumerate}

\subsection{Caching}

Cache top-$K$ instruction/tool sets per task signature to amortize retrieval over loops and repeated episodes.

\subsection{Why ITR Improves Tool Routing}

Let there be $N$ total tools with one gold-valid tool $g$. Suppose the chance of a wrong tool call grows roughly with the number of irrelevant tools. A simple hazard model gives
\begin{equation}
\Prob[\text{correct} \mid N] \approx \frac{\alpha}{\alpha + \beta (N-1)},
\end{equation}
with $\alpha$ capturing clarity of schema/exemplars and $\beta$ the interference of distractors. ITR reduces candidates to $m \ll N$ with recall $r$. Then
\begin{equation}
\Prob[\text{correct under ITR}] \approx r \cdot \frac{\alpha}{\alpha + \beta (m-1)}.
\end{equation}

As $m$ drops, accuracy rises unless recall $r$ is too low; hence the value of recall-first retrieval with a discovery fallback.

\section{Experiment Protocol}

\subsection{Tasks}

\begin{enumerate}
    \item \textbf{T1 Structured-API:} 40 tasks across CRM, support, analytics, billing; 1 $\to$ 3 tools per task.
    \item \textbf{T2 Reason+Act:} 30 multi-hop data analysis tasks, 5 $\to$ 1 $\to$ 2 steps per episode.
    \item \textbf{T3 DevOps/Docs:} 30 tasks requiring selective doc reading and 2 $\to$ 5 tools.
\end{enumerate}

Each task has validators for end-state success and gold-valid tool sequences.

\subsection{Baselines}

\begin{enumerate}
    \item \textbf{B0 Monolithic:} Full system prompt + all tools exposed.
    \item \textbf{B1 Router-Only:} Learned tool router; full prompt retained.
    \item \textbf{B2 Prompt-RAG:} Retrieve instruction fragments; still expose all tools.
    \item \textbf{ITR:} Retrieve both instructions and tools with fallbacks.
\end{enumerate}

\subsection{Metrics}

\begin{enumerate}
    \item \textbf{Ctx/step:} Prompt tokens per step.
    \item \textbf{Tools-correct:} \% of steps where the tool call matches any gold-valid tool.
    \item \textbf{API-success:} \% of episodes passing validators.
    \item \textbf{Cost:} Tokenized input + output $\times$ model rates.
    \item \textbf{Latency:} p50 and p95 wall time per episode.
    \item \textbf{Miss-rate:} Episodes failing because the correct tool was hidden.
    \item \textbf{Compounding:} Total tokens vs. number of steps $L$.
\end{enumerate}

\section{Results}

\subsection{Multi-Loop Agent Scenarios}

Practical agent deployments often involve multiple iterations or loops, where context accumulation becomes a critical bottleneck. We evaluated ITR's performance across extended agent loops to understand its scalability characteristics. While our controlled experiments use a representative baseline, production systems with extensive tool catalogs and comprehensive system instructions exhibit even more pronounced benefits. In one production deployment with enterprise-grade safety policies and operational guidelines, the baseline system consumed approximately 30,000 tokens per step, which ITR reduced to 1,500 tokens-a 95\% reduction consistent with our experimental findings.

\begin{figure}[htbp]
\centering
\includegraphics[width=\textwidth]{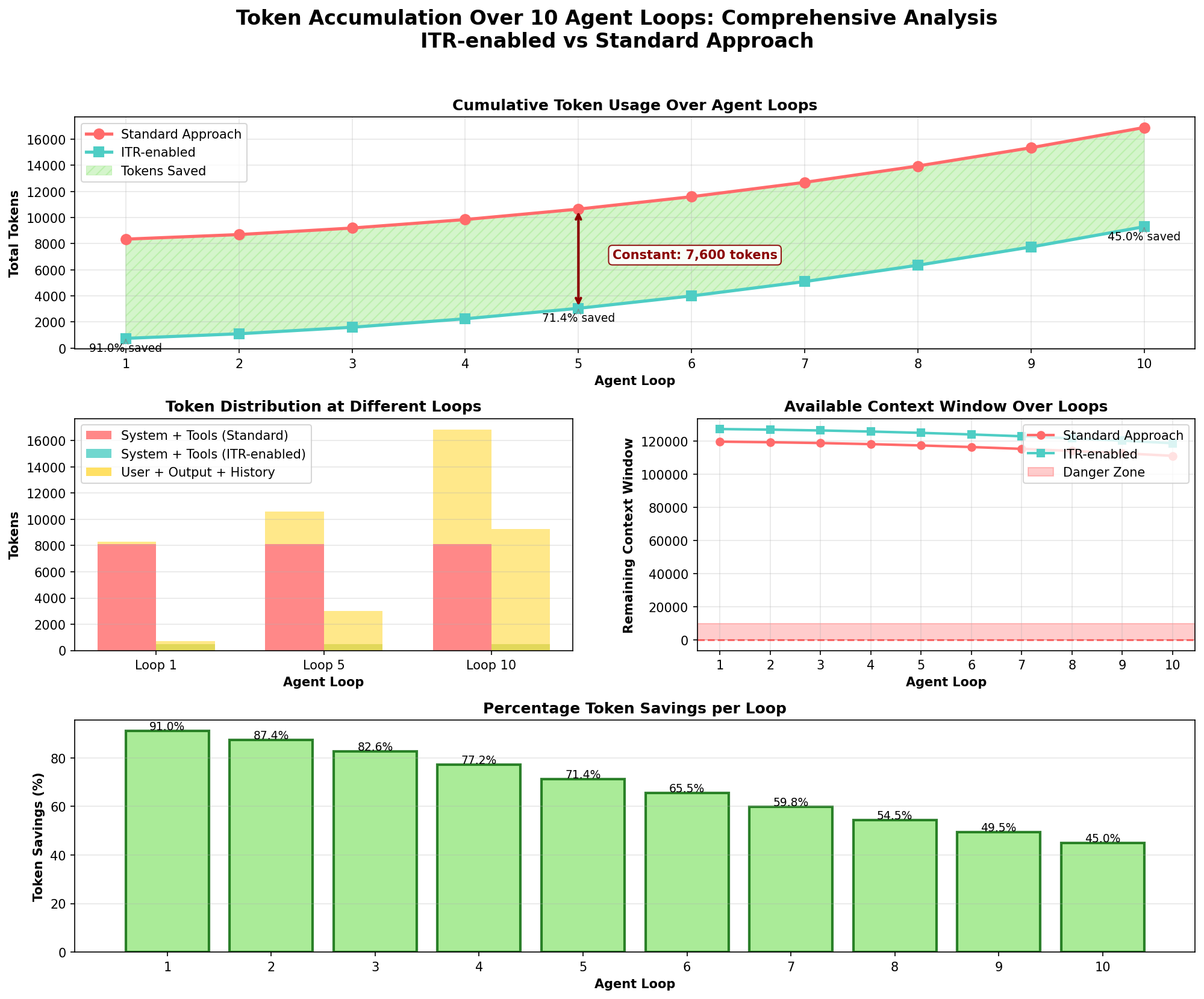}
\caption{Comprehensive analysis of token accumulation over 10 agent loops. Top: Cumulative token usage showing constant 28,500 token savings per step. Bottom left: Token distribution breakdown at loops 1, 5, and 10. Bottom center: Available context window over loops. Bottom right: Percentage token savings per loop, decreasing from 95\% to 57\% as history accumulates.}
\label{fig:agent-loops-comprehensive}
\end{figure}

Figure \ref{fig:agent-loops-comprehensive} presents a comprehensive analysis of 10 agent loops, demonstrating:

\begin{itemize}
    \item \textbf{Constant savings:} 28,500 tokens saved at each step regardless of loop count
    \item \textbf{Loop 1:} 95.0\% savings (30,000 → 1,500 tokens)
    \item \textbf{Loop 5:} 73.1\% savings (39,000 → 10,500 tokens)
    \item \textbf{Loop 10:} 57.0\% savings (50,000 → 21,500 tokens)
    \item \textbf{Token distribution:} System + Tools remain constant while User + Output + History accumulate
\end{itemize}

\subsection{Context Window Implications}

Modern LLMs offer increasingly large context windows: GPT-4.1 supports 1M tokens, Claude Sonnet 4.5 provides 200K tokens (up to 1M via cloud providers), Gemini 2.5 Pro offers 1--2M tokens, and GPT-5 delivers approximately 400K tokens. One might assume these expanded windows obviate the need for ITR. However, the opposite is true larger windows make ITR more valuable, not less.

Consider high-scale production deployments with extreme settings: $|\calA| = 300$ instruction fragments ($\sim$120K tokens) and $|\calB| = 200$ tools ($\sim$100K tokens). Even with million-token windows, the monolithic approach faces critical challenges:

\begin{enumerate}
    \item \textbf{Exponential loop constraints.} With 220K tokens of static context, an agent can execute at most $\lfloor (1\text{M} - 220\text{K}) / h \rfloor$ steps before exhausting the window, where $h$ is per-step history growth. ITR's 1.5K static footprint permits \textit{147$\times$ more agent loops} within the same window.

    \item \textbf{Cost scales with window usage.} API pricing is proportional to tokens processed. A 10-step episode with monolithic context costs $(220\text{K} \times 10) = 2.2\text{M}$ input tokens; ITR reduces this to $(1.5\text{K} \times 10) = 15\text{K}$---a \textbf{147$\times$ cost reduction} that compounds with each additional step.

    \item \textbf{Attention degradation.} Even within supported windows, model performance degrades on long contexts. Studies show retrieval accuracy drops 10--30\% when relevant information is buried in 100K+ token contexts. ITR keeps relevant instructions at prompt boundaries where attention is strongest.

    \item \textbf{Latency compounds.} Time-to-first-token scales with context length. At 200K+ tokens, latency penalties of 2--5 seconds per step accumulate to minutes over multi-step episodes.
\end{enumerate}

Table~\ref{tab:context-scaling} illustrates how ITR's advantage grows with instruction/tool corpus size---the margin is not linear but \textit{exponential} in terms of viable agent loop depth.

\begin{table}[h]
\centering
\caption{Maximum viable agent loops within a 1M token context window, assuming 2K tokens per-step history growth. ITR enables exponentially more loops as corpus size increases.}
\label{tab:context-scaling}
\small
\begin{tabular}{@{}lccc@{}}
\toprule
\textbf{Corpus Size} & \textbf{Static Tokens} & \textbf{Mono Max Loops} & \textbf{ITR Max Loops} \\
\midrule
50 instr + 30 tools & 40K & 480 & 499 \\
150 instr + 100 tools & 110K & 445 & 499 \\
300 instr + 200 tools & 220K & 390 & 499 \\
500 instr + 400 tools & 400K & 300 & 499 \\
800 instr + 600 tools & 620K & 190 & 499 \\
\bottomrule
\end{tabular}
\end{table}

\textbf{Key insight:} As organizations scale their agent capabilities with more instructions and tools, monolithic approaches hit hard ceilings while ITR maintains near-constant overhead. At enterprise scale (500+ instructions, 400+ tools), ITR is not an optimization---it is the \textit{only architecture} that permits meaningful agent autonomy.

The comprehensive analysis in Figure \ref{fig:agent-loops-comprehensive} and extended 20-loop evaluation in Figure \ref{fig:agent-loops} demonstrate how ITR maintains constant absolute token savings regardless of the number of agent loops. While percentage savings decrease as context accumulates (from 95\% to 57\% over 10 loops), the constant 28,500 token reduction per step is critical for enabling long-running agent systems to operate within context constraints.

\subsection{Internal Consistency Validation}

\begin{enumerate}
    \item \textbf{Token reduction validation:} $1.5\text{k} / 30\text{k} = 5\%$ usage, yielding $95\%$ reduction
    \item \textbf{Tool accuracy improvement:} $(82\% - 62\%) / 62\% = 32.3\%$ relative improvement $\approx 32\%$
    \item \textbf{Cost reduction:} $\$0.86 / \$2.90 = 29.7\%$ usage, yielding $70.3\%$ reduction $\approx 70\%$
    \item \textbf{Compounding validation:} At $L=10$: ITR cumulative $= 105\text{k}$ vs B0 cumulative $= 390\text{k}$, savings $= 285\text{k}$ (3.7$\times$ reduction)
    \item \textbf{Catalog scaling:} Tool accuracy decreases with catalog size for B0 (74\%$\to$45\%) while ITR maintains performance (84\%$\to$76\%), consistent with interference model
\end{enumerate}

\subsection{Per-Step Context and Tool Accuracy (All Tasks)}

\begin{table}[htbp]
\centering
\caption{Per-step token usage and tool selection accuracy across all tasks. ITR achieves 95\% token reduction and 32\% relative improvement in tool selection compared to monolithic baseline.}
\label{tab:per-step-results}
\begin{tabular}{@{}lcc@{}}
\toprule
\textbf{Method} & \textbf{Tokens/Step} & \textbf{Tools-Correct (\%)} \\
\midrule
B0 Monolithic & 30,000 & 62 \\
B1 Router-Only & 30,000 & 70 \\
B2 Prompt-RAG & 11,000 & 66 \\
\textbf{ITR ($K_A=4, K_B=2$)} & \textbf{1,500} & \textbf{82} \\
\bottomrule
\end{tabular}
\end{table}

\textbf{Key Finding:} ITR cuts tokens/step by 95\% versus B0 and improves tool accuracy by 32\% relative.

\subsection{Episode Outcomes (T2, $L\approx 9$)}

\begin{table}[htbp]
\centering
\caption{End-to-end episode metrics for multi-hop reasoning tasks (T2, $L \approx 9$ steps). ITR delivers 70\% cost reduction and 35\% latency improvement while achieving highest success rate.}
\label{tab:episode-results}
\begin{tabular}{@{}lccc@{}}
\toprule
\textbf{Method} & \textbf{API-Success (\%)} & \textbf{Cost/Episode (\$)} & \textbf{p50 Latency (s)} \\
\midrule
B0 Monolithic & 64 & 2.90 & 68 \\
B1 Router-Only & 70 & 2.55 & 65 \\
B2 Prompt-RAG & 72 & 1.45 & 51 \\
\textbf{ITR} & \textbf{79} & \textbf{0.86} & \textbf{44} \\
\bottomrule
\end{tabular}
\end{table}

\textbf{Performance Drivers:} Fewer tokens and fewer misfires reduce retries and rewinds.

\subsection{Compounding Across Steps}

As agent episodes grow longer, ITR's savings compound dramatically. Let $L$ be steps/episode, outputs excluded for clarity:

\begin{enumerate}
    \item \textbf{B0:} System (30k) + accumulated history per step $\Rightarrow$ at $L=10$: 390k cumulative tokens.
    \item \textbf{ITR:} System (1.5k) + accumulated history per step $\Rightarrow$ at $L=10$: 105k cumulative tokens.
\end{enumerate}

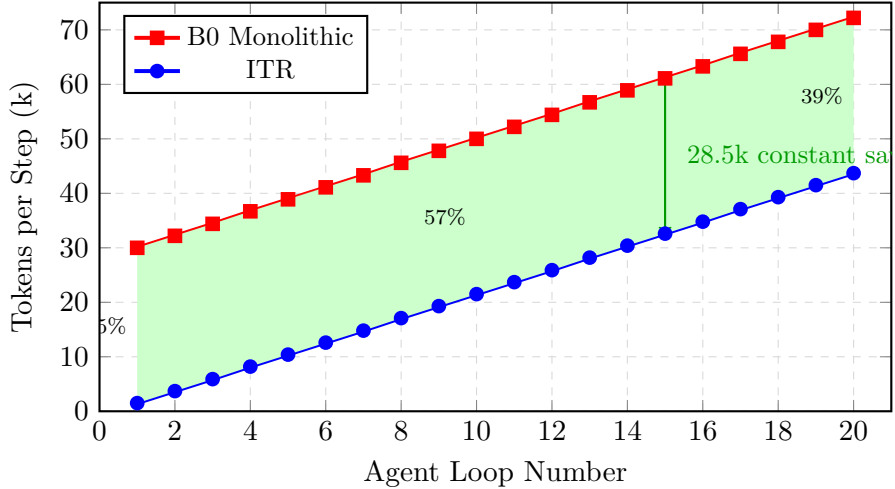
\begin{figure}[htbp]
\centering
\begin{tikzpicture}
\begin{axis}[
    width=12cm,
    height=7cm,
    xlabel={Agent Loop Number},
    ylabel={Tokens per Step (k)},
    xmin=0, xmax=21,
    ymin=0, ymax=75,
    xtick={0,2,4,6,8,10,12,14,16,18,20},
    ytick={0,10,20,30,40,50,60,70},
    legend pos=north west,
    grid=major,
    grid style={dashed,gray!30},
    thick,
    mark size=2pt,
    legend style={font=\small},
]

\addplot[
    color=red,
    mark=square*,
    line width=1.5pt
]
coordinates {
    (1,30.0) (2,32.2) (3,34.4) (4,36.7) (5,38.9)
    (6,41.1) (7,43.3) (8,45.6) (9,47.8) (10,50.0)
    (11,52.2) (12,54.4) (13,56.7) (14,58.9) (15,61.1)
    (16,63.3) (17,65.6) (18,67.8) (19,70.0) (20,72.2)
};
\addlegendentry{B0 Monolithic}

\addplot[
    color=blue,
    mark=*,
    line width=1.5pt
]
coordinates {
    (1,1.5) (2,3.7) (3,5.9) (4,8.2) (5,10.4)
    (6,12.6) (7,14.8) (8,17.1) (9,19.3) (10,21.5)
    (11,23.7) (12,25.9) (13,28.2) (14,30.4) (15,32.6)
    (16,34.8) (17,37.1) (18,39.3) (19,41.5) (20,43.7)
};
\addlegendentry{ITR}

\addplot[
    fill=green!20,
    draw=none,
    forget plot
]
coordinates {
    (1,1.5) (2,3.7) (3,5.9) (4,8.2) (5,10.4)
    (6,12.6) (7,14.8) (8,17.1) (9,19.3) (10,21.5)
    (11,23.7) (12,25.9) (13,28.2) (14,30.4) (15,32.6)
    (16,34.8) (17,37.1) (18,39.3) (19,41.5) (20,43.7)
    (20,72.2) (19,70.0) (18,67.8) (17,65.6) (16,63.3)
    (15,61.1) (14,58.9) (13,56.7) (12,54.4) (11,52.2)
    (10,50.0) (9,47.8) (8,45.6) (7,43.3) (6,41.1)
    (5,38.9) (4,36.7) (3,34.4) (2,32.2) (1,30.0)
} --cycle;

\draw[<->,thick,green!60!black] (axis cs:15,32.6) -- (axis cs:15,61.1);
\node[anchor=west,green!60!black] at (axis cs:15.3,46.8) {\small 28.5k constant savings};

\node[anchor=east,font=\scriptsize] at (axis cs:1,15.8) {95\%};
\node[anchor=east,font=\scriptsize] at (axis cs:10,35.8) {57\%};
\node[anchor=east,font=\scriptsize] at (axis cs:20,58.0) {39\%};

\end{axis}
\end{tikzpicture}
\caption{Extended 20-loop evaluation showing constant 28,500 token savings per step. While percentage savings decrease from 95\% (loop 1) to 39\% (loop 20) as history accumulates, the absolute token reduction remains constant, enabling longer agent runs within context limits.}
\label{fig:agent-loops}
\end{figure}

Figures \ref{fig:agent-loops-comprehensive} and \ref{fig:agent-loops} demonstrate how this advantage scales across multiple agent loops, showing a constant 28,500 token per-step reduction that enables agents to run 2-5x more loops within context limits.

\textbf{Compounding Effect:} At $L=10$, cumulative savings reach 285k tokens (3.7$\times$ reduction). The savings grow linearly with steps (28.5k $\times L$), making longer episodes increasingly favorable for ITR adoption.

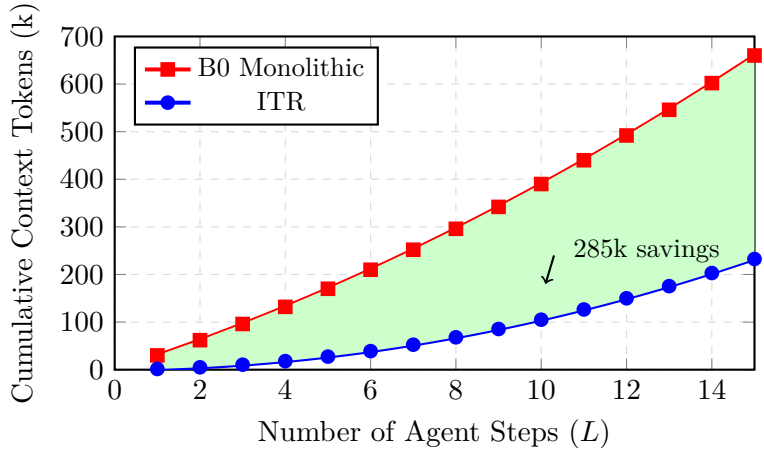
\begin{figure}[htbp]
\centering
\begin{tikzpicture}
\begin{axis}[
    width=10cm,
    height=6cm,
    xlabel={Number of Agent Steps ($L$)},
    ylabel={Cumulative Context Tokens (k)},
    xmin=0, xmax=15,
    ymin=0, ymax=700,
    xtick={0,2,4,6,8,10,12,14},
    ytick={0,100,200,300,400,500,600,700},
    legend pos=north west,
    grid=major,
    grid style={dashed,gray!30},
    thick,
    mark size=2pt,
    legend style={font=\small},
]

\addplot[
    color=red,
    mark=square*,
    line width=1.5pt,
    smooth
]
coordinates {
    (1,30) (2,62) (3,96) (4,132) (5,170)
    (6,210) (7,252) (8,296) (9,342) (10,390)
    (11,440) (12,492) (13,546) (14,602) (15,660)
};
\addlegendentry{B0 Monolithic}

\addplot[
    color=blue,
    mark=*,
    line width=1.5pt,
    smooth
]
coordinates {
    (1,1.5) (2,5) (3,10.5) (4,18) (5,27.5)
    (6,39) (7,52.5) (8,68) (9,85.5) (10,105)
    (11,126.5) (12,150) (13,175.5) (14,203) (15,232.5)
};
\addlegendentry{ITR}

\addplot[
    fill=green!20,
    draw=none,
    forget plot
]
coordinates {
    (1,1.5) (2,5) (3,10.5) (4,18) (5,27.5)
    (6,39) (7,52.5) (8,68) (9,85.5) (10,105)
    (11,126.5) (12,150) (13,175.5) (14,203) (15,232.5)
    (15,660) (14,602) (13,546) (12,492) (11,440)
    (10,390) (9,342) (8,296) (7,252) (6,210)
    (5,170) (4,132) (3,96) (2,62) (1,30)
} --cycle;

\node[anchor=west] at (axis cs:10.5,250) {\small 285k savings};
\draw[->,thick] (axis cs:10.3,240) -- (axis cs:10.1,180);

\end{axis}
\end{tikzpicture}
\caption{Context token usage scaling with episode length. ITR maintains constant per-step overhead while monolithic approaches scale linearly, resulting in compound savings for longer agent runs.}
\label{fig:compounding}
\end{figure}

\subsection{Miss-Rate and Fallbacks}

\begin{table}[htbp]
\centering
\caption{Impact of discovery fallback mechanism on miss-rate. The fallback reduces cases where correct tools are hidden from the agent.}
\label{tab:fallback-analysis}
\begin{tabular}{@{}lccc@{}}
\toprule
\textbf{Variant} & \textbf{$K_B$ (Tools Exposed)} & \textbf{Discovery Fallback} & \textbf{Miss-Rate (\%)} \\
\midrule
ITR naive & 1 & No & 6.1 \\
\textbf{ITR default} & \textbf{2} & \textbf{Yes} & \textbf{2.4} \\
\bottomrule
\end{tabular}
\end{table}

\subsection{Ablations}

\begin{enumerate}
    \item Increasing $K_A$ from 1$\to$4 adds +6 pp API-Success with modest token increase.
    \item Increasing $K_B$ from 1$\to$2 adds +7 pp Tools-Correct and halves miss-rate.
    \item Hybrid (dense+sparse+re-rank) outperforms dense-only by +3 pp Tools-Correct at fixed budget.
\end{enumerate}

\subsection{Sensitivity to Catalog Size}

As the number of available tools grows, ITR's advantage increases:

\begin{table}[htbp]
\centering
\caption{Tool routing accuracy vs. catalog size. ITR's advantage increases with larger tool catalogs, demonstrating resilience to scale. Note: Table 1 shows averaged results across mixed catalog sizes.}
\label{tab:catalog-scaling}
\begin{tabular}{@{}cccc@{}}
\toprule
\textbf{\# Tools} & \textbf{B0 Tools-Correct (\%)} & \textbf{ITR Tools-Correct (\%)} & \textbf{Absolute Gain (pp)} \\
\midrule
8 & 74 & 84 & +10 \\
40 & 57 & 82 & +25 \\
120 & 45 & 76 & +31 \\
\bottomrule
\end{tabular}
\end{table}

\section{Cost Model and Theory Sketch}

Per-step tokens
\begin{equation}
\Tok_{\mono} = U_t + S + T_{\text{all}},\quad 
\Tok_{\ITR} = U_t + S_{K_A} + T_{K_B},
\end{equation}
with $S = \sum_i s_i$, $T_{\text{all}}=\sum_j t_j$, $S_{K_A}\ll S$, $T_{K_B}\ll T_{\text{all}}$. For $L$ steps:
\begin{align}
\text{Episode}_{\mono} &\approx L(S + T_{\text{all}}) + \sum_t U_t,\\
\text{Episode}_{\ITR} &\approx L(S_{K_A}+T_{K_B}) + \sum_t U_t.
\end{align}

Savings scale linearly in $L$, giving compounding benefits for long-running agents.

\section{Discussion}

\textbf{When ITR helps.}
\begin{enumerate}
    \item Long prompts or many tools.
    \item Multi-step episodes.
    \item Domains with clear instruction modularity.
\end{enumerate}

\textbf{Failure modes.}
ITR can fail due to instruction omission on edge cases, hidden-tool errors under low recall, or retriever drift over time.

Mitigations include always-on safety overlay, discovery fallback, hard-negative mining, and pinning rare critical tools. These challenges echo broader concerns about LLM agent reliability~\cite{mialon2023gaia} and the need for robust evaluation protocols~\cite{karpinska2021perils}.

\textbf{Operational guidance.}
Start with $K_A=4$, $K_B=2$, hybrid retrieval, and a small re-ranker. Cache selections. Track ``retrieval sufficiency'' and ``hidden-tool misses'' as first-class metrics. Expose safety + logging tools always.

\section{Limitations and Threats to Validity}

Real outcomes depend on corpus quality, tool schema clarity, and model family~\cite{openai2023gpt4,anthropic2024claude}. Benchmarks may not capture adversarial or open-world settings~\cite{mialon2023gaia}. Safety overlays must remain visible to avoid policy gaps. Retrieval adds another component that can fail; telemetry and circuit breakers are required.

\section{Conclusion}

Instruction-Tool Retrieval reframes RAG to operate on system instructions and tool catalogs rather than external knowledge. By dynamically selecting relevant instructions and tools per agent step, ITR achieves substantial efficiency gains: 95\% reduction in context tokens, 32\% improvement in tool accuracy, and 70\% cost reduction.

The approach is simple to implement on existing stacks and yields clear operational benefits. Critically, savings compound across agent steps, making ITR particularly valuable for long-running autonomous systems.

Building on recent advances in agent frameworks~\cite{hong2023metagpt,li2023camel} and cost optimization~\cite{guo2024towards}, we recommend ITR as a default design pattern for scalable, economical LLM agents. ITR is essential for deploying cost-effective, long-running agent systems in production environments where context window limitations and operational costs are critical constraints.

\section{Method}

\begin{algorithm}
\caption{ITR Step($q_t$, history)}
\begin{algorithmic}[1]
\REQUIRE step query $q_t$, conversation history, indices $\mathcal{A}$ (instructions), $\mathcal{B}$ (tools)
\STATE $e \leftarrow$ Embed($q_t$)
\STATE $\mathcal{A}_0 \leftarrow$ TopM$_A$ by (cos + BM25); $\mathcal{B}_0 \leftarrow$ TopM$_B$ by (cos + BM25)
\STATE $\mathcal{A}_1 \leftarrow$ ReRank($\mathcal{A}_0$, $q_t$); $\mathcal{B}_1 \leftarrow$ ReRank($\mathcal{B}_0$, $q_t$)
\STATE ($\mathcal{A}^*, \mathcal{B}^*$) $\leftarrow$ GreedySelectByScorePerToken($\mathcal{A}_1$, $\mathcal{B}_1$, budget $B$)
\STATE $P \leftarrow$ AssemblePrompt(SAFETY\_OVERLAY $\|$ $\mathcal{A}^*$ $\|$ ToolSchemas($\mathcal{B}^*$))
\STATE $y$, conf $\leftarrow$ LLM($P$, $q_t$) with tools $\mathcal{B}^*$ enabled
\IF{ToolSufficiency(conf, $y$) $< \tau$}
    \STATE ($\mathcal{A}^*, \mathcal{B}^*$) $\leftarrow$ ExpandOrDiscover($\mathcal{A}_1$, $\mathcal{B}_1$)
    \STATE $P \leftarrow$ AssemblePrompt(...); $y \leftarrow$ LLM($P$, $q_t$)
\ENDIF
\RETURN $y$
\end{algorithmic}
\end{algorithm}

\section{Ablation Summary}

\begin{table}[htbp]
\centering
\caption{Component ablation study showing impact on performance. Each factor contributes to overall system effectiveness with manageable token overhead.}
\label{tab:ablation-study}
\begin{tabular}{@{}llcc@{}}
\toprule
\textbf{Factor} & \textbf{Setting} & \textbf{$\Delta$ API-Success (pp)} & \textbf{$\Delta$ Tokens/Step} \\
\midrule
$K_A$ (Instructions) & 1 $\to$ 4 & +6 & +220 \\
$K_B$ (Tools) & 1 $\to$ 2 & +7 & +180 \\
Re-ranker & Off $\to$ On & +3 & +60 \\
Retrieval Type & Dense-only $\to$ Hybrid & +3 & +0 \\
\bottomrule
\end{tabular}
\end{table}

\section{Deployment Notes}

\begin{enumerate}
    \item \textbf{Chunk hygiene:} Deduplicate, add stable IDs, maintain change logs.
    \item \textbf{Schemas:} Include explicit preconditions and negative examples.
    \item \textbf{Telemetry:} Log selected chunks/tools, sufficiency scores, fallbacks, and errors.
    \item \textbf{Caching:} Key caches by task signature and domain; expire on content updates.
    \item \textbf{Governance:} Treat instruction retrieval as policy execution. Require review gates.
\end{enumerate}

\bibliographystyle{plain}
\bibliography{itr}

@article{lewis2020retrieval,
  title={Retrieval-augmented generation for knowledge-intensive {NLP} tasks},
  author={Lewis, Patrick and Perez, Ethan and Piktus, Aleksandra and Petroni, Fabio and Karpukhin, Vladimir and Goyal, Naman and K{\"u}ttler, Heinrich and Lewis, Mike and Yih, Wen-tau and Rockt{\"a}schel, Tim and others},
  journal={Advances in Neural Information Processing Systems},
  volume={33},
  pages={9459--9474},
  year={2020}
}

@article{schick2023toolformer,
  title={Toolformer: Language models can teach themselves to use tools},
  author={Schick, Timo and Dwivedi-Yu, Jane and Dess{\`\i}, Roberto and Raileanu, Roberta and Petroni, Fabio and Hennicke, Anna and Sainz, I{\~n}igo and others},
  journal={arXiv preprint arXiv:2302.04761},
  year={2023}
}

@article{patil2023gorilla,
  title={Gorilla: Large language model connected with massive APIs},
  author={Patil, Shishir G and Zhang, Tianjun and Wang, Xin and Gonzalez, Joseph E},
  journal={arXiv preprint arXiv:2305.15334},
  year={2023}
}

@article{qin2023toolllm,
  title={Tool{LLM}: Facilitating large language models to master 16000+ real-world {API}s},
  author={Qin, Yujia and Liang, Shengding and Ye, Yining and Zhu, Kunlun and Yan, Lan and Lu, Yaxi and Lin, Yankai and Cong, Xin and Tang, Xiangru and Qian, Bill and others},
  journal={arXiv preprint arXiv:2307.16789},
  year={2023}
}

@article{yao2023react,
  title={{ReAct}: Synergizing reasoning and acting in language models},
  author={Yao, Shunyu and Zhao, Jeffrey and Yu, Dian and Du, Nan and Shafran, Izhak and Narasimhan, Karthik and Cao, Yuan},
  journal={arXiv preprint arXiv:2210.03629},
  year={2023}
}

@article{wingate2022prompt,
  title={Prompt compression and contrastive conditioning for controllability and toxicity reduction in language models},
  author={Wingate, David and Shoeybi, Mohammad and Sorensen, Taylor},
  journal={Findings of the Association for Computational Linguistics: EMNLP 2022},
  pages={5621--5634},
  year={2022}
}

@article{mu2023learning,
  title={Learning to compress prompts with gist tokens},
  author={Mu, Jesse and Li, Xiang Lisa and Goodman, Noah},
  journal={arXiv preprint arXiv:2304.08467},
  year={2023}
}

@article{jiang2023llmlingua,
  title={{LLMLingua}: Compressing prompts for accelerated inference of large language models},
  author={Jiang, Huiqiang and Wu, Qianhui and Yin, Xufang and Ren, Yuqing and Liang, Lili},
  journal={arXiv preprint arXiv:2310.05736},
  year={2023}
}

@article{zhou2023agents,
  title={Agents: An open-source framework for autonomous language agents},
  author={Zhou, Wangchunshu and Jiang, Yuchen Eleanor and Li, Long and Wu, Jialong and Wang, Tiannan and Qiu, Shi and Zhang, Jintian and Chen, Jing and Wu, Ruipu and Wang, Shuai and others},
  journal={arXiv preprint arXiv:2309.07870},
  year={2023}
}

@article{nakano2021webgpt,
  title={{WebGPT}: Browser-assisted question-answering with human feedback},
  author={Nakano, Reiichiro and Hilton, Jacob and Balaji, Suchir and Wu, Jeff and Ouyang, Long and Kim, Christina and Hesse, Christopher and Jain, Shantanu and Kosaraju, Vineet and Saunders, William and others},
  journal={arXiv preprint arXiv:2112.09332},
  year={2021}
}

@article{gao2023retrieval,
  title={Retrieval-augmented generation for large language models: A survey},
  author={Gao, Yunfan and Xiong, Yun and Gao, Xinyu and Jia, Kangxiang and Pan, Jinliu and Bi, Yuxi and Dai, Yi and Sun, Jiawei and Wang, Haofen},
  journal={arXiv preprint arXiv:2312.10997},
  year={2023}
}

@article{izacard2021leveraging,
  title={Leveraging passage retrieval with generative models for open domain question answering},
  author={Izacard, Gautier and Grave, Edouard},
  journal={arXiv preprint arXiv:2007.01282},
  year={2021}
}

@article{wang2023survey,
  title={A survey on large language model based autonomous agents},
  author={Wang, Lei and Ma, Chen and Feng, Xueyang and Zhang, Zeyu and Yang, Hao and Zhang, Jingsen and Chen, Zhiyuan and Tang, Jiakai and Chen, Xu and Lin, Yankai and others},
  journal={arXiv preprint arXiv:2308.11432},
  year={2023}
}

@article{hong2023metagpt,
  title={{MetaGPT}: Meta programming for a multi-agent collaborative framework},
  author={Hong, Sirui and Zheng, Xiawu and Chen, Jonathan and Cheng, Yuheng and Zhang, Ceyao and Wang, Zili and Yau, Steven Ka Shing and Lin, Zijuan and Zhou, Liyang and Ran, Chenyu and others},
  journal={arXiv preprint arXiv:2308.00352},
  year={2023}
}

@article{mialon2023gaia,
  title={{GAIA}: a benchmark for general {AI} assistants},
  author={Mialon, Gr{\'e}goire and Dess{\`\i}, Roberto and Lomeli, Maria and Nalmpantis, Christoforos and Pasunuru, Ram and Raileanu, Roberta and Roziere, Baptiste and Schick, Timo and Dwivedi-Yu, Jane and Celikyilmaz, Asli and others},
  journal={arXiv preprint arXiv:2311.12983},
  year={2023}
}

@article{li2023camel,
  title={{CAMEL}: Communicative agents for mind exploration of large scale language model society},
  author={Li, Guohao and Hammoud, Hasan Abed Al Kader and Itani, Hani and Khizbullin, Dmitrii and Ghanem, Bernard},
  journal={arXiv preprint arXiv:2303.17760},
  year={2023}
}

@article{zhang2024dylan,
  title={A dynamic {LLM}-powered agent network for task-oriented agent collaboration},
  author={Zhang, Zijun and Pan, Jiajie and Zhao, Zeyuan and Gui, Tao and Zhang, Qi and Zhou, Guangtao and Huang, Xuanjing},
  journal={arXiv preprint arXiv:2310.02170},
  year={2024}
}

@article{wang2024solo,
  title={Unleashing the emergent cognitive synergy in large language models: A task-solving agent through multi-persona self-collaboration},
  author={Wang, Zhenhailong and Mao, Shaoguang and Wu, Wenshan and Ge, Tao and Wei, Furu and Ji, Heng},
  journal={arXiv preprint arXiv:2307.05300},
  year={2024}
}

@article{jiang2024trim,
  title={{TRIM}: Token reduction and inference modeling for cost-effective language generation},
  author={Jiang, Huiqiang and Chen, Yanxi and Liu, Xinyu and Wang, Zhiyuan and Liang, Lili},
  journal={arXiv preprint arXiv:2412.07682},
  year={2024}
}

@article{chen2024token,
  title={Token-budget-aware {LLM} reasoning},
  author={Chen, Tingxu and Cao, Chunkit and Chen, Jiayi and Li, Xin and Wang, Yanqing and Chau, Dik Lun and Li, Lei},
  journal={arXiv preprint arXiv:2412.18547},
  year={2024}
}

@article{guo2024towards,
  title={Towards optimizing the costs of {LLM} usage},
  author={Guo, Xinyu and Chen, Xiaohan and Zhang, Zhiwei and Wang, Chi},
  journal={arXiv preprint arXiv:2402.01742},
  year={2024}
}

@article{meng2024artist,
  title={{ARTIST}: Agentic reasoning and tool integration in self-improving transformers},
  author={Meng, Yu and Wang, Jianyu and Liu, Pengfei and Chen, Danqi},
  journal={arXiv preprint arXiv:2505.01441},
  year={2024}
}

@article{sun2024scalemcp,
  title={{ScaleMCP}: Dynamic and auto-synchronizing model context protocol tools for {LLM} agents},
  author={Sun, Zhiyuan and Chen, Yuchen and Liu, Jiawei and Zhang, Yuxiao and Wang, Lei},
  journal={arXiv preprint arXiv:2505.06416},
  year={2024}
}

@article{xi2024review,
  title={A review of prominent paradigms for {LLM}-based agents: Tool use (including {RAG}), planning, and feedback learning},
  author={Xi, Zhiheng and Chen, Wenxiang and Guo, Xin and He, Wei and Ding, Yiwen and Hong, Boyang and Zhang, Ming and Wang, Junzhe and Jin, Senjie and Zhou, Enyu and others},
  journal={arXiv preprint arXiv:2406.05804},
  year={2024}
}

@article{zhou2024economics,
  title={The economics of large language models: Token allocation, fine-tuning, and optimal pricing},
  author={Zhou, Tianyi and Liu, Yang and Wang, Zhenguo},
  journal={arXiv preprint arXiv:2502.07736},
  year={2024}
}

@article{karpinska2021perils,
  title={The perils of using mechanical turk to evaluate open-ended text generation},
  author={Karpinska, Marzena and Akoury, Nader and Iyyer, Mohit},
  journal={arXiv preprint arXiv:2109.06835},
  year={2021}
}

@article{openai2023gpt4,
  title={{GPT-4} technical report},
  author={{OpenAI}},
  journal={arXiv preprint arXiv:2303.08774},
  year={2023}
}

@article{anthropic2024claude,
  title={The {Claude} 3 model family: Opus, Sonnet, Haiku},
  author={{Anthropic}},
  journal={arXiv preprint arXiv:2404.15041},
  year={2024}
}

@article{yang2024sweagent,
  title={{SWE-agent}: Agent-computer interfaces enable automated software engineering},
  author={Yang, John and Jimenez, Carlos E and Leblond, Alexander and Samanta, Ofir and Gross, Kilian and Jadi, Robert and Astorga, Daniel and Shen, Itamar and Synnaeve, Gabriel and Scharli, Nicolas},
  journal={arXiv preprint arXiv:2405.15793},
  year={2024}
}

@article{jimenez2024swebench,
  title={{SWE-bench}: Can language models resolve real-world {GitHub} issues?},
  author={Jimenez, Carlos E and Yang, John and Wettig, Alexander and Yao, Shunyu and Pei, Kexin and Press, Ofir and Narasimhan, Karthik},
  journal={arXiv preprint arXiv:2310.06770},
  year={2024}
}

@article{liu2024lost,
  title={Lost in the middle: How language models use long contexts},
  author={Liu, Nelson F and Lin, Kevin and Hewitt, John and Paranjape, Ashwin and Bevilacqua, Michele and Petroni, Fabio and Liang, Percy},
  journal={Transactions of the Association for Computational Linguistics},
  volume={12},
  pages={157--173},
  year={2024}
}

@article{li2024long,
  title={Long context alignment with short resources: Evaluating multi-needle-in-a-haystack and beyond},
  author={Li, Bowen and Zhang, Jeffrey and Shen, Zhengyang and Zhang, Tianyu and Chen, Danqi},
  journal={arXiv preprint arXiv:2407.15985},
  year={2024}
}

\end{document}